%% file: aaai22.tex
\def\year{2022}\relax
\documentclass[letterpaper]{article} 
\usepackage{aaai22}  
\usepackage{times}  
\usepackage{helvet}  
\usepackage{courier}  
\usepackage[hyphens]{url}  
\usepackage{graphicx} 
\usepackage{natbib}  
\usepackage{caption} 
\DeclareCaptionStyle{ruled}{labelfont=normalfont,labelsep=colon,strut=off} 
\usepackage{algorithm}

\usepackage{newfloat}
\usepackage{listings}
\usepackage{booktabs}       
\usepackage{xcolor}         
\usepackage{amsmath}
\usepackage{algpseudocode} 
\DeclareMathOperator*{\argminB}{\emph{argmin}}

\usepackage{cleveref}
\usepackage{multirow}
\algnewcommand{\algorithmicor}{\textbf{ or }}
\algnewcommand{\OR}{\algorithmicor}
\usepackage{capt-of}
\usepackage{subcaption}

\newcommand{\yc}[1]{\textcolor{black}{#1}}
\newcommand{\ycao}[1]{\textcolor{black}{#1}}
\newcommand{\gk}[1]{\textcolor{black}{#1}}
\newcommand{\jb}[1]{\textcolor{black}{#1}}

\floatstyle{ruled}
\newfloat{listing}{tb}{lst}{}
\floatname{listing}{Listing}
\title{Gradient-based Novelty Detection  \\Boosted by Self-supervised Binary Classification}
\author {
    Jingbo Sun\textsuperscript{\rm 1}, 
    Li Yang\textsuperscript{\rm 1},  
    Jiaxin Zhang\textsuperscript{\rm 2}, 
    Frank Liu\textsuperscript{\rm 2}, 
    Mahantesh Halappanavar\textsuperscript{\rm 3}, \\
    Deliang Fan\textsuperscript{\rm 1}, 
    Yu Cao\textsuperscript{\rm 1}
}
\affiliations {
    \textsuperscript{\rm 1} Arizona State University\\
    \textsuperscript{\rm 2} Oak Ridge National Laboratory\\
    \textsuperscript{\rm 3} Pacific Northwest National Laboratory\\
    jsun127@asu.edu, lyang166@asu.edu, zhangj@ornl.gov, liufy@ornl.gov,  Mahantesh.Halappanavar@pnnl.gov, dfan@asu.edu, Yu.Cao@asu.edu
}

\usepackage{bibentry}

\begin{document}

\maketitle

\begin{abstract}
Novelty detection aims to automatically identify out-of-distribution (OOD) data, without any prior knowledge of them. It is a critical step in data monitoring, behavior analysis and other applications, helping enable continual learning in the field. Conventional methods of OOD detection perform multi-variate analysis on an ensemble of data or features, and usually resort to the supervision with OOD data to improve the accuracy. In reality, such supervision is impractical as one cannot anticipate the anomalous data. 
In this paper, we propose a novel, self-supervised approach that does not rely on any pre-defined OOD data: (1) The new method evaluates the Mahalanobis distance of the gradients between the in-distribution and OOD data. (2) It is assisted by a self-supervised binary classifier to guide the label selection to generate the gradients, and maximize the Mahalanobis distance. 
In the evaluation with multiple datasets, such as CIFAR-10, CIFAR-100, SVHN and TinyImageNet, the proposed approach consistently outperforms state-of-the-art supervised and unsupervised methods in the area under the receiver operating characteristic (AUROC) and area under the precision-recall curve (AUPR) metrics. We further demonstrate that this detector is able to accurately learn one OOD class in continual learning.
\end{abstract}

\input{introduction}

\input{methdology}
\input{pseudocode}
\input{experiment}

\input{ablation}

\input{conclusion}


\bibliography{aaai22}

\end{document}

%% file: introduction.tex
\section{Introduction}

Deep neural networks (DNNs) have achieved high accuracy in many fields, such as image classification, natural language processing, and speech recognition. Their success is built upon carefully handcrafted DNN architectures, big data collection and expensive model training. A well-trained model promises high inference accuracy if the input falls into the distribution of the training data. However, in many real-world scenarios, there is no guarantee that the input is always in the distribution. The encounter with out-of-distribution (OOD) input is inevitable due to the difficulty in data collection, unforeseeable user scenarios, and complex dynamics. 

To manage the emergence of OOD data at the first moment, it is vitally important to have an accurate novelty detector that continuously evaluates the data stream and alarms the system once OOD data arrives. Upon the detection of OOD arrival, the system can then manage the situation with three possible methods: (1) It can rely on the OOD detector to collect new data and send them back to the data center, such that OOD data can be combined with previous in-distribution data (IDD) to re-train the model; (2) It can temporally utilize the detector as a one-class classifier to recognize the new class of OOD, in addition to existing IDD classes; and (3) It can activate a continual learning method at the edge to adapt the model in the field, such as the multi-head method with the assistance of the OOD detector. In all three cases, the accuracy and robustness of the novelty detector guard the success of continual model adaptation and knowledge update.

\begin{figure}[!t]
\centering
\includegraphics[width=0.45\textwidth]{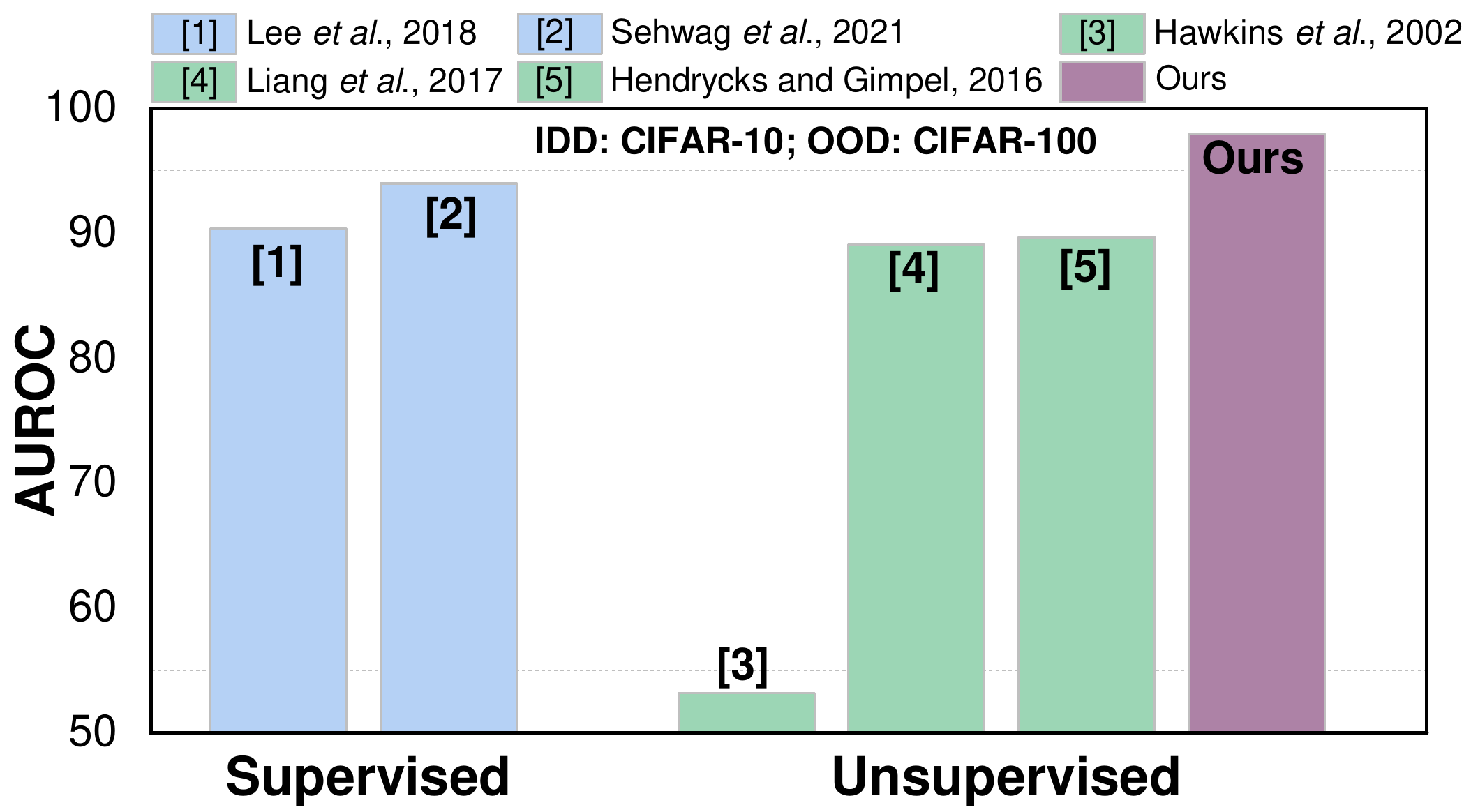}
\caption{Comparison of AUROC between our proposed method and other state-of-the-art methods.}
\label{fig:aurocintro}
\vspace{-3mm}
\end{figure}

In this work, we propose a self-supervised approach which generates a set of OOD data from unsupervised statistical analysis, instead of supervised labeling. This set of OOD data is then combined with IDD to train a binary classifier, which in turn, helps boost the performance of the unsupervised detector that collects more OOD data for training. As this mutual process continues, our novelty detector achieves higher and higher confidence in OOD detection. The contributions of this paper are as follows:

\begin{itemize}
 \item Gradient-based novelty detection that employs the Mahalanobis distance as the metric to differentiate in-distribution and OOD input. The gradients are generated from a pre-trained classifier for IDD only, without any pre-knowledge of OOD. 
 \item A self-supervised binary classifier. As previous works demonstrated, the availability of a binary classifier helps boost the accuracy. Yet distinguished from them, we don't rely on any labelled OOD data to train the classifier. The training set is initialized by the gradient-based detector. In turn, the binary classifier pre-screens OOD and IDD, and guides the selection of labels in the gradient-based detector to maximize the distance calculation. Through such mutual assistance, our approach is unsupervised and continually improves the accuracy.
 \item High accuracy in OOD detection and one-class classification. We evaluate our methodology in a comprehensive set of benchmarks. 
 As shown in Fig.~\ref{fig:aurocintro}, our self-supervised method consistently achieves higher AUROC than other supervised and unsupervised results, confirming the advantages in gradient-based novelty detection.
 \end{itemize}

%% file: methdology.tex
\section{Background}

Current OOD detectors usually use the information extracted from either the data itself or features projected by the feedforward path in the IDD engine. ~\cite{hendrycks2016baseline} 
demonstrated that the softmax score of the outlier tended to be higher compared with IDD and thus, thresholding inputs based on this score is feasible to detect the OOD. ~\cite{liang2017enhancing} improved this idea by introducing input preprocessing and temperature scaling. ~\cite{lee2018simple} proved that the in-distribution samples formed a multivariate Gaussian distribution in the high dimensional feature space, and proposed to use the Mahalanobis distance to measure how far an outlier is away from this in-distribution. There also exist autoencoder based OOD detectors ~\cite{abati2019latent,chen2017outlier,hawkins2002outlier,kwon2020novelty,sakurada2014anomaly,zhou2017anomaly} 
which use the reconstruction loss from the decoder to characterize the novelty. 
These data or activation-based methods demonstrated the value in OOD detection. On the other side, they have not explored one important step in the development of DNNs, the gradients back-propagated from the classification layer. These gradients present the first-order derivative to adapt the model and improve the separation of multiple classes. They contain a rich body of information to differentiate OOD from IDD.

To collect the gradients from the classifier that is prepared for the IDD, a label is required for cross-entropy loss and back-propagation. However, one challenge in the gradient-based approach is that labels of OOD data are not available in the process of novelty detection. To address this issue, a recent work by ~\cite{lee2020gradients} introduced the confounding label, which only triggers small gradients for the IDD input. 
The gradients of the OOD input would be larger since they introduce many new features that are different from IDD. ~\cite{kwon2020novelty} explored the gradient-based representations of the novelty but they avoid the label issue by 
proposing a directional gradient constraint to the loss function so that the gradient direction of the OOD input does not align with the ones of the IDD. However, this method requires re-training of the model. 
In contrast, in this work, we utilize the gradient-based approach and the pre-trained model without any modification. In addition, we only use the training labels to collect the gradient.

Note that to boost the accuracy in novelty detection, many prior approaches utilize supervised training. They adopted a small amount of OOD samples to pre-train the novelty detector. For example, ~\cite{lee2018simple} trained a logistic regression detector to estimate the weights for feature ensemble. ~\cite{lee2020gradients} trained a binary detector to distinguish the gradient representation of the in-distribution and OOD input. However, this type of supervised training requires the availability of labelled OOD data up front, which restricts its application in reality.

\section{Methodology}

\begin{figure*}[!t]
\centering
\includegraphics[width=\textwidth]{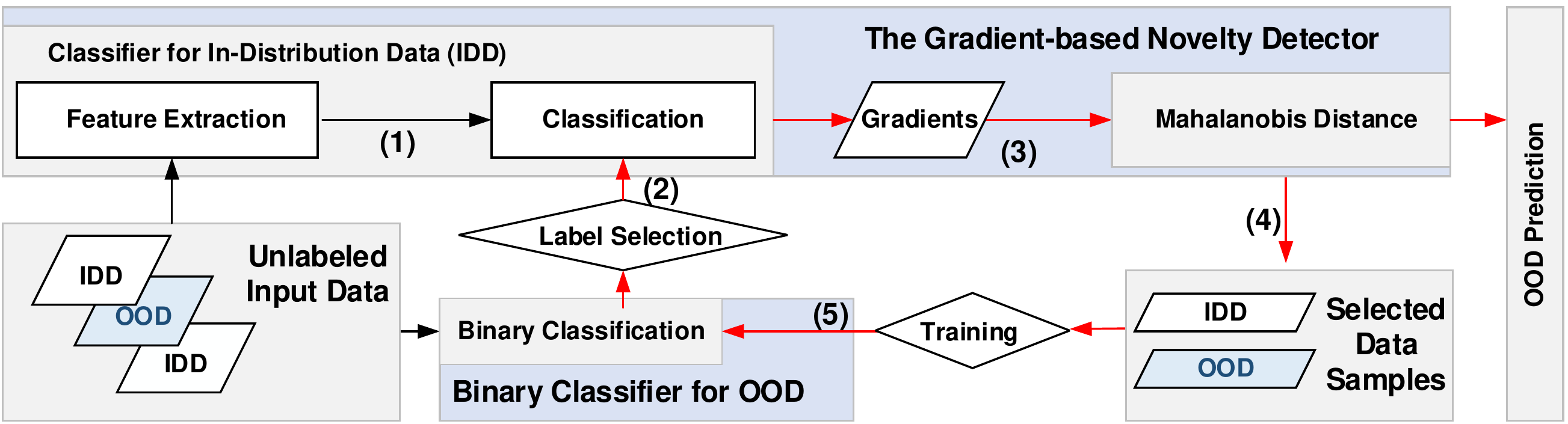}
\caption{The flow of our self-supervised OOD detector. (1) A DNN-based classifier is first prepared for IDD data. (2) The binary classifier predicts IDD or OOD. If the result is OOD, then in the IDD classifier, it selects the appropriate label to maximize the distance. (3) Based on the selected label, gradients are generated by the IDD classifier for the calculation of Mahalanobis distance. (4) From each patch of data, a balanced set of IDD and OOD data is selected based on the Mahalanobis distance. (5) This balanced IDD/OOD data set is used to continuously train the binary classifier. }
\vspace{-2mm}
\label{fig:framework}
\end{figure*}

The overarching goal of our methodology is to accurately identify OOD data from IDD. A successful OOD detection is equivalent to correctly classify the OOD input as one new class (i.e., one-class classification). For IDD inputs, they will be classified to the previous known classes. To achieve this goal, we propose a closed-loop methodology that interleaves the unsupervised ODD detector based on the Mahalanobis distance, with a binary IDD/OOD classifier. 
Fig.~\ref{fig:framework} illustrates our methodology, consisting of these two main components:

\begin{itemize}
    \item A gradient-based novelty detector: Distinguished from previous works that analyzed either the input data or the features, our detector exploits the gradients propagated backward from the classification layer for statistical analysis. The classifier in this step is designed for IDD only and all labels are from the IDD classes. To maximize the Mahalanobis distance between OOD and IDD classes, we select the appropriate class to label the OOD data, which is predicted by a binary classifier, as in Fig.\ref{fig:framework}.
    \item A self-supervised binary classifier. This classifier is designed to pre-screen IDD and OOD data, in order to assist label selection to generate the gradients. While the structure of this binary classifier is similar as that in \cite{goodfellow2014generative}, the training does not rely on any labeled data, but from a balanced set that is selected by the detector based on the Mahalanobis distance. 
\end{itemize}

Overall, the mutual assistance between these two components helps accomplish OOD detection without the need of external supervision. In the following subsections, we first introduce our main processing path in OOD detection: 
the Mahalanobis distance in the gradient space. We then introduce 
the binary classifier as a necessary assistant to boost the performance of the main path, as well as
the self-supervised training of it.
Finally, we describe how these two components mutually assist each other with a thorough case study.

\subsection{Gradient-based novelty detector}

 
Our approach starts from a given dataset of in-distribution data (IDD), and a deep neural network (DNN) based classifier trained by this IDD dataset. Note that at this step, only labels for IDD classes are accessible. In our examples of the image classification task, previous works have demonstrated that such a DNN
 is capable of 
 separating the manifolds of each in-distribution class and achieve high classification accuracy. After the training of the DNN is completed, the gradients of each in-distribution sample, which is back-propagated from its ground-truth label, is distributed within a small range around the manifold of its predicted class,
 forming class-specific distributions that correspond to each individual manifold.
 
When an OOD data is given to this pre-trained DNN, it will lie far away from any manifold of IDD. Without any knowledge about this OOD data, if we still perform gradient-based training supervised by the IDD label, the DNN model will experience much larger gradients which force the model to reconstruct itself toward the OOD data.
 In this context, the distribution of gradients will generate different characteristics for IDD and OOD, paving a promising path toward novelty detection. Therefore we propose to utilize the Mahalanobis distance in the gradient space as the statistical metric to separate the outliers from the in-distribution ones.


The Mahalanobis distance 
is defined by the following: 
\begin{equation}
M_{x} =  ({\nabla_{c}}f(x)-\hat{\mu}_{c})^T  \hat{\Sigma}^{-1}({\nabla_{c}}f(x)- \hat{\mu}_{c})
\label{fig:Mahalanobis}
\end{equation}

where $\hat{\mu}_{c}$ is the gradient mean of the in-distribution samples $\mathcal{X}_{in}$ of the class $c$ and $\hat{\Sigma}^{-1}$ is the tied precision matrix of all the known(training) classes. We use the equations below to estimate these two parameters:

\begin{equation}
\hat{\mu}_{c} = \frac{1}{N_{c}}\sum\limits_{i:y_{i}=c} {\nabla_{c}}f({x^{in}_{i}})
\label{fig:mean}
\end{equation}

\begin{equation}
\hat{\Sigma}=  \frac{1}{N}\sum\limits_{c}\sum\limits_{i:y_{i}=c}({\nabla_{c}}f({x^{in}_{i}})-\hat{\mu}_{c} )({\nabla_{c}}f({x^{in}_{i}})-\hat{\mu}_{c})^\top
\label{fig:tied}
\end{equation}
where $N_{c}$ is the amount of the IDD samples in the class $c$, $N$ is the amount of the entire IDD training dataset.

We use Equations (\ref{fig:mean}) and (\ref{fig:tied}) to characterize the gradient distribution of the IDD by using the same training dataset of the DNN. After this class-specific distributions estimation is done, we use equation (\ref{fig:Mahalanobis}) to measure how the gradient of the new input deviates from these estimated distributions.

In Equation (\ref{fig:Mahalanobis}), ${\nabla_{c}}f(x)$ is the back-propagated gradient with respect to the class $c$. This raises a question of how to calculate the gradient of the OOD input since their ground-truth label 
$Y_{ood}$ is not in the IDD label space $\mathcal{Y}_{in}$. To solve this problem, we propose to use the predicted label $Y_{ood}^{Pred} \in  \mathcal{Y}_{in}$ to calculate and cross-entropy loss $\mathcal{L}(Y_{ood}^{Pred}; X_{ood}; \Theta)$ and do the back-propagation. From the manifold perspective, this means that we select the manifold closest to the input sample. This manifold requires the minimal amount of adaptation to the new input, i.e., the minimal gradient and Mahalanobis distance. If such Mahalanobis distance is still large, the input sample has a high probability to be an outlier.

\begin{figure}[!t]
\hspace{-2mm}
\includegraphics[width=0.51\textwidth]{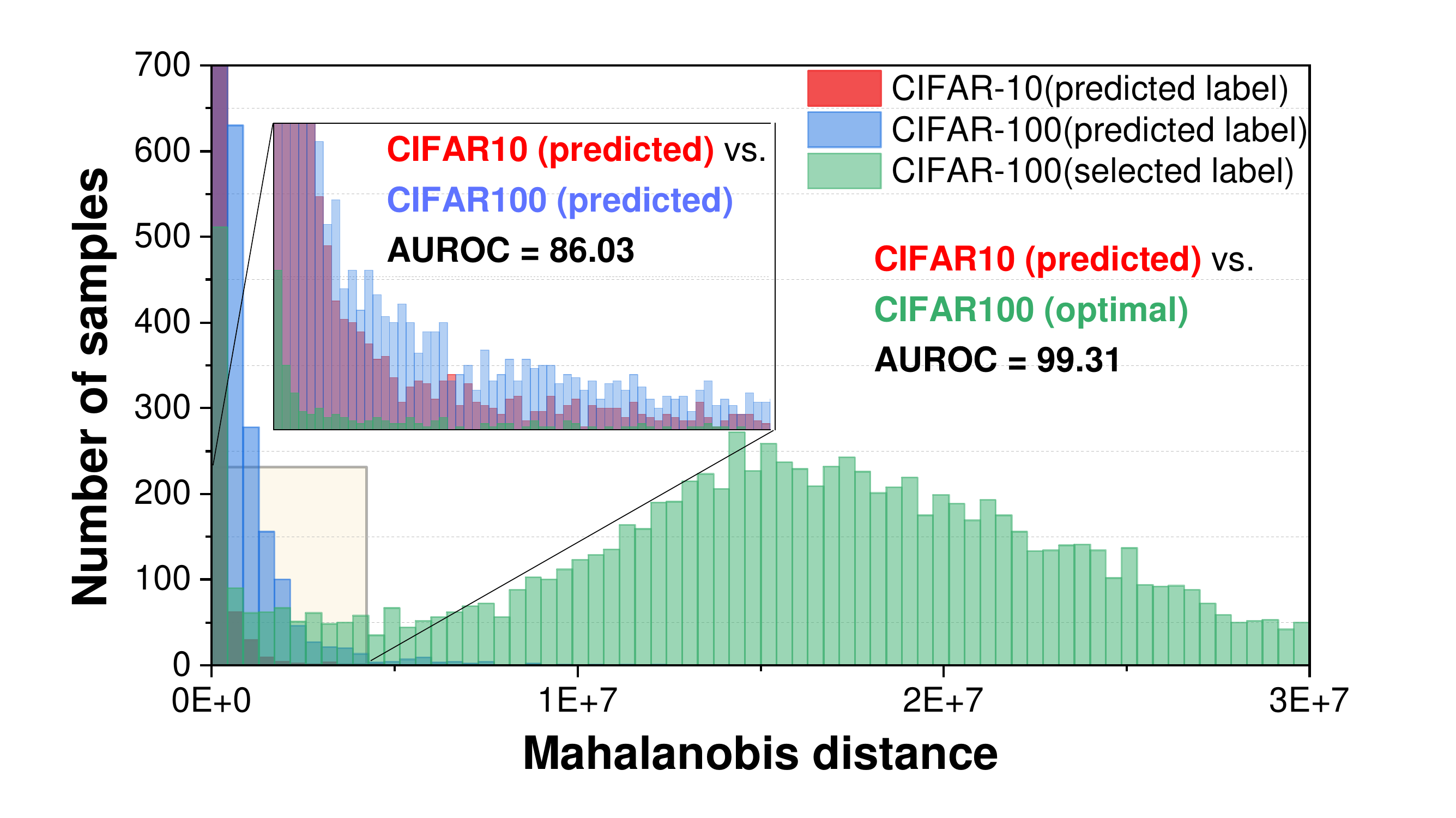}

\vspace{-3mm}
\caption{The distribution of the Mahalanobis distance of in-distribution data using predicted label (highlighted in red), and OOD using both predicted label (highlighted in blue) and selected label (highlighted in green). By using the selected label for all the outliers, their novelty score distribution shifts further away farther away from blue to green, as shown in the main figure, with less overlap with IDD data (the red distribution, as shown in the inset.)}
\label{fig:hist}
\end{figure}

\begin{table}[t]
\centering
\resizebox{0.48\textwidth}{!}{
\begin{tabular}{c|ccc}
\toprule
\begin{tabular}[c]{@{}c@{}}IDD (CIFAR-10)\end{tabular} & \multicolumn{3}{c}{AUROC}                                                                                          \\ \midrule 
 OOD                                      & \multicolumn{1}{c}{\begin{tabular}[c]{@{}c@{}}State-of-\\the-art\end{tabular}} & \multicolumn{1}{c}{\begin{tabular}[c]{@{}c@{}}Ours (with \\predicted label)\end{tabular}} & \begin{tabular}[c]{@{}c@{}}Ours (with \\selected label)\end{tabular} \\ \midrule
TinyImageNet                    & \multicolumn{1}{c}{99.50}            & \multicolumn{1}{c}{91.10}      & \textbf{99.92}                    \\ 
SVHN                         & \multicolumn{1}{c}{99.90}            & \multicolumn{1}{c}{91.63}          & \textbf{99.99}                      \\ 
LSUN                         & \multicolumn{1}{c}{99.70}            & \multicolumn{1}{c}{90.08}          & \textbf{99.99}                    \\ 
CIFAR-100                      & \multicolumn{1}{c}{93.40}            & \multicolumn{1}{c}{86.03}      & \textbf{97.99}                    \\ \bottomrule
\end{tabular}
}

\caption{Comparison of AUROC results between the state-of-the-art~\protect\cite{lee2018simple,sehwag2021ssd} and our proposed novelty detector.
The middle column is the performance of our novelty detector using the predicted label from the DNN for IDD to calculate the loss, gradients and Mahalanobis distance. The right column shows the performance boosting if we intentionally select the label to maximize the distance between IDD and OOD.}
\label{table1}
\end{table}

As shown in Table~\ref{table1}, if only using the predicted label by the IDD classifier, our proposed method is not competitive to the state-of-the-art method. This is due to the high overlap of the novelty score distribution between the IDD and OOD. 
\yc{In fact,} the Mahalanobis distance is minimal if we use the predicted label as the ground truth label for back-propagation. \yc{To overcome this barrier,} 
\yc{we} intentionally \yc{select} a different label to maximize the Mahalanobis distance for all the OOD samples. By doing that, we expect to \yc{achieve} a larger mean of the novelty score distribution for the OOD. \yc{Consequently, it will be easier to} threshold these OOD samples \yc{to reach} higher detection accuracy.

Fig.~\ref{fig:hist} illustrates the novelty score distribution of the IDD using the predicted label and OOD using both predicted and a pre-selected label. 
After intentionally using the selected label for all the OOD samples, the corresponding score distributions shift away from the in-distribution \yc{and thus,} significantly improves the AUROC result (Table~\ref{table1}, Column 3). However, this raises a new problem: how to \yc{trigger the usage of different labels for IDD and OOD (i.e., predicted by the IDD classifier and the selected label, respectively), in the calculation of gradients?}
\yc{Here we introduce} a binary classifier to pre-screen the input and make the initial IDD/OOD prediction. Based on the prediction result, our novelty detector chooses different label for gradient calculation.

 \textbf{Label selection:} To select a label $y^{Opt}$ that gives the maximum gradient \yc{distance}, we propose to use the predicted softmax class probability:
 
\begin{equation}
\\y^{Custom} = \argminB_c(\sum\limits_{i}Softmax(f(x_i^{OOD};\Theta))) 
\label{fig:optimallabel}
\end{equation}

We use Equation (\ref{fig:optimallabel}) to select the class that has the minimal average softmax probability for a batch of OOD data. This is equivalent to finding the least likely class for an OOD data to fall into. By using $y^{Opt}$, its cross-entropy loss $\mathcal{L}(y^{Opt}; X_{ood}; \Theta)$ becomes \yc{the} maximum which results in the largest gradient and Mahalanobis distance. The only question left is where to find the OOD samples to estimate this label. We will address this issue in subsection \ref{labelselection}.


\subsection{Self-supervised binary classifier}
 \label{labelselection} 
 To guide our novelty detector to use either the predicted or selected label in case of in-distribution or OOD input, we introduce a simple self-supervised binary classifier \yc{to screen IDD and OOD. The output from this binary classifier will be used to guide label selection in the pre-trained IDD classifier for gradient generation}: 
 \begin{itemize}
     \item For the predicted IDD input, the pre-trained model uses the predicted label $y^{Pred}$ for back-propagation. 
     \item For the predicted OOD input, the model uses the selected label $y^{Opt}$. 
 \end{itemize}

 \textbf{\yc{Unsupervised preparation of OOD samples for initial training}:} One key feature of our proposed method is to be self-supervised, which means no OOD sample is available in advance. Therefore to create \yc{a dataset for the training of the} binary classifier, 
 we utilize the predicted IDD/OOD samples selected by the gradient-based novelty detector. For example, \yc{assuming} the binary classifier is randomly initialized and a batch $X^1$ mixed with IDD and OOD data comes in, our novelty detector will first calculate the novelty confidence score $S^1$ = \{$s_1, s_2, ...,s_N$\}$_1$ using the predicted label. We select N/2 samples that correspond to the highest and lowest confidence scores in the $S^1$ as the binary classifier training data set $X^{pred} = \{X_{in}^{pred}, X_{ood}^{pred}\}$ . This step helps select the best possible in-distribution and OOD data from the current batch so that the binary classifier's training inputs are reasonable. We use the $X_{ood}^{pred}$ to select the label
 and use the $X^{pred}$ to train the binary classifier. Once training is done, the following input batches will involve the cooperation from both the gradient-based novelty detector and the binary classifier.

\subsection{Mutual assistance \yc{between the binary classifier and the Mahalanobis path}}

The entire system is continuously exposed to a stream of unlabeled mini batch $X^1,X^2$,..., where each $X^{i}$ consists of N samples \{$x_1, x_2, ...,x_N$\} mixed with IDD and OOD data. Due to the small size of the first batch $X^1$ and the relatively low performance of the novelty detector using the predicted label, the initial training of the binary classifier could not guarantee to be success. Therefore, an enhanced training (Fig.~\ref{fig:framework} and Algorithm 1) is required when more data is available. The training routine consists of three major steps: (1) \yc{Initial prediction of the} binary classifier;
(2) \yc{Calculation of the} Mahalanobis score;
and (3) \yc{Re-training of the} binary classifier. 

\textbf{\yc{Initial prediction of the} binary classifier 
:} When the new batch $X^{k}$ = \{$x_1, x_2, ...,x_N$\}$_k$ \yc{arrives,}
the system first concatenates this new batch $X^{k}$ with all the previously stored batches $X^{k-1}... X^{1}$. Given this concatenated batch $X^{all}$, the binary classifier makes its initial outlier prediction $\hat{Y}$= \{$y'_1, y'_2, ...,y'_{N\times k} | y'_i\in(0,1)$\}. 

\textbf{\yc{Calculation of the} Mahalanobis score:} The gradient-based novelty detector takes the prediction result from the binary classifier and the concatenated batch. For each sample $x_{i}$ in batch, our novelty detector first classifies it to one of the in-distribution class $y^{Pred}_{i}\in  \mathcal{Y}_{in}$ , then checks the binary classifier prediction $y'_{i}$. If $y'_{i}$ is 0 (predicted in-distribution), use $y^{Pred}_{i}$ as the ground truth label in the loss function for back-propagation, otherwise, use the pre-selected label $y^{Opt}$. After the gradient is available, we calculate the Mahalanobis distance $s_{i}$ as the novelty confident score. 

\textbf{\yc{Re-training of the} binary classifier:} Given novelty confident score $S^{all}$ =\{$s_1, s_2, ...,s_{N\times k}$\} from the gradient-based novelty detector, we select $(N\times k)/2$ samples from the concatenated batch that correspond to the highest and lowest scores in $S^{all}$ as a new training dataset for the binary classifier. Before training, we re-initialize the binary classifier to make sure the previous model will not be inherited into the current stage. Once training is done, our system is updated with the knowledge of all previous batches and is ready to process next available inputs with higher detection accuracy. 

\yc{As shown in Fig.~\ref{fig:framework}, this mutual assistance continues with more unlabelled data, which keeps improving the accuracy of both the unsupervised detection engine and the binary classifier in this closed loop.}

%% file: pseudocode.tex
\algnewcommand\algorithmicforeach{\textbf{for each}}
\algdef{S}[FOR]{ForEach}[1]{\algorithmicforeach\ #1\ \algorithmicdo}

\begin{algorithm}[!t]
    \label{fig:alg1}
	\caption{Gradient-based Novelty Detection Boosted by Self-supervised Binary Classification
} 
        \textbf{Input:} In-distribution gradient distributions \{${\hat\mu}, {\hat\Sigma}^{-1}$\}, batches for testing $[X^{1},X^{2},...,X^{k}] $

	\begin{algorithmic}[1]
        
        \Function{$noveltyScore$ }{$X,{\hat\mu},{\hat\Sigma}^{-1},\hat{Y},y^{Opt}$}
        \State $Mahalanobis\_Distance = list()$
        \ForEach {$x\;in\;X$}  
        \State $c$ = $y^{Opt} \;\;\;\;\;// Label\;for\;back$-$propagation$
        \If {$\hat{Y}$ is None\OR Binary Classifier predict x as IDD}    
\State $c$ = Novelty detector predicted label  
        \EndIf
        \State $Score = ({\nabla_{c}}f(x)-\hat{\mu}_{c})^T  \hat{\Sigma}^{-1}({\nabla_{c}}f(x)- \hat{\mu}_{c})$
        \State $Mahalanobis\_Distance.append(Score)$
        \EndFor
        \State \Return $Mahalanobis\_Distance$
        \EndFunction

\State ${X^{all}}\;=\;None$
	    \While {new batch ${X^k}$ is available}
\State ${X^{all}}={X^{all}} + {X^{k}} \;\;\;\;//batch\; concatenation$
	    \If{${X^{k}}$\;is\;the\;first\;batch}
	    
\State ${S^{all}}=noveltyScore( {X^{all}},{\hat\mu},{\hat\Sigma}^{-1})$ 
	    \Else
\State $\hat{Y}= binary\;classifier\;prediction\;on\;X^{all}$
\State ${S^{all}}= noveltyScore( {X^{all}},{\hat\mu},{\hat\Sigma}^{-1},\hat{Y},y^{Opt})$
	    \EndIf
\State Based\;on\;${S^{all}}$, select\;samples\;from\;${X^{all}}$\; \\
with\;the\;highest/lowest\;score\;as\;predicted\;$X_{in}/X_{ood}$
	    \State Use \{$X_{in},X_{ood}$\} to train binary classifier
	    \If {$y^{Opt}\;is\;None$}
\State $y^{Opt}= \underset{c}{\operatorname{argmin}}(\sum\limits_{i}Softmax(f(x_{i}^{ood})))$
	    \EndIf
	    \EndWhile
	\State\textbf{return}  $X_{i}\in X_{all}$ is an outlier if $ S^{all}_i > threshold$
	\end{algorithmic} 

\end{algorithm}
\vspace{-2mm}

%% file: experiment.tex
\section{Experiments}


We use three pre-trained ResNet-34 networks ~\cite{resnet} provided by~\cite{lee2018simple} as \yc{the base of} our gradient-based novelty detector. Each model is trained on CIFAR-10, CIFAR-100~\cite{cifar} and SVHN~\cite{netzer2011reading} with the testing accuracy of 93.67\%, 78.34\% and 96.68\% accordingly. To calculate the gradient-based Mahalanobis distance, we only use the gradient extracted from the last layer of the feature extractor. Our simple binary classifier has the structure of three convolution layers and one batch normalization layer with a Sigmoid classifier. It is trained by minimizing the cross-entropy loss using Adam~\cite{kingma2014adam}. The initial learning rate is set to 0.0002 and the decay rate is controlled by $\beta_{1} $ = 0.5 and $\beta_{2} $ = 0.999. We train it for 500 epochs in both initial training and re-training process. 
Regarding the batch size, we find that the amount of IDD and OOD samples in each batch, and the batch size itself have a strong impact on the performance of the binary classifier and the overall system, as discussed in subsection~\ref{batchsize}.

Our experiment includes three in-distribution datasets: CIFAR-10, CIFAR-100 and SVHN. We test each of them using the other two as the OOD. \yc{In addition}, we use two more OOD datasets: the resized version of the ImageNet~\cite{ILSVRC15} and LSUN~\cite{yu2016lsun} provided by~\cite{liang2017enhancing}. Each of these two datasets contains 10,000 images with size 32 by 32. For all three in-distribution datasets, we use only the testing portion of the images because the training portion has already been used to train the novelty detector. For each IDD/OOD pair, we randomly select 5,000/5,000 (IDD/OOD) as the training dataset and further divide them into mini batches with size of 100/100 (IDD/OOD) to emulate the data streaming, the rest 5,000/5,000 (IDD/OOD) samples are used to test the binary classifier accuracy and the overall novelty detection performance after each new batch has been taken into the system. We evaluate our detector with two performance metrics: AUROC and AUPR. All experiments are performed with PyTorch~\cite{paszke2017automatic} on one NVIDIA GeForce RTX 2080 platform.

\input{auroctable2}

\subsection{\yc{Batch training of the} binary classifier}
\label{batchsize}
Inspired from the discriminator in the GAN \cite{goodfellow2014generative} that can successfully distinguish the real images from the fake ones, we use the similar discriminator loss in our binary classifier training, where the loss from the IDD and OOD data are calculated separately. During the prediction phase, the binary classifier achieves high accuracy if the input batch contains either in-distribution or OOD samples. \yc{This is because the batch norm layer averages the difference within each group of data (i.e., IDD or OOD) while stressing the difference between IDD and OOD inputs.} 
For each novelty detection experiment, we \yc{adopt this batch-based approach and} present the results of using different batch size where each batch contains only IDD or OOD data.

\Cref{AUPRtable,aupr} present the performance of our proposed method
\yc{as compared to} other novelty detectors. 
Our method outperforms all previous supervised and unsupervised works across all IDD/OOD setup. In particular, our method improves the AUROC of the experiment where CIFAR100 as IDD and CIFAR10 as OOD by up to $\sim$5 and $\sim$7 with batch size 32 and 128 accordingly. These two datasets are very similar \yc{to each other.} Therefore it's extremely \yc{challenging} to detect the outlier \yc{in previous approaches.} \yc{With the new framework,} our method \yc{significantly improves the state-of-the-art.}

Fig.~\ref{fig:landscape} illustrates the stepwise improvement of our framework. Each point in the curve corresponds to the intermediate testing accuracy of the binary classifier and the AUROC of the system after every new batch \yc{of unlabelled data} (100/100) arrives. From the curves, we can \yc{observe} the \yc{efficacy of} mutual assistance between the binary classifier and the novelty detector. The initial testing accuracy of the binary classifier reaches around 90 in all three IDD/OOD experiments, \yc{which} proves that our self-supervised approach based on the novelty detector's output is \yc{effective}. As more data \yc{is received}, every re-training on the binary classifier improves its accuracy, \yc{contributed by increasingly higher confidence of} the IDD/OOD prediction from the novelty detector. In \yc{turn}, the novelty detector's performance is boosted, benefiting from \yc{higher accuracy} of the binary classifier \yc{and better label selection. Such positive feedback eventually drives the performance of the overall framework, reaching high accuracy of both the novelty detector and the binary classifier.} 
\begin{figure}[!ht]
\centering
\begin{subfigure}[t]{0.236\textwidth}
    \centering
    \includegraphics[width=\textwidth]{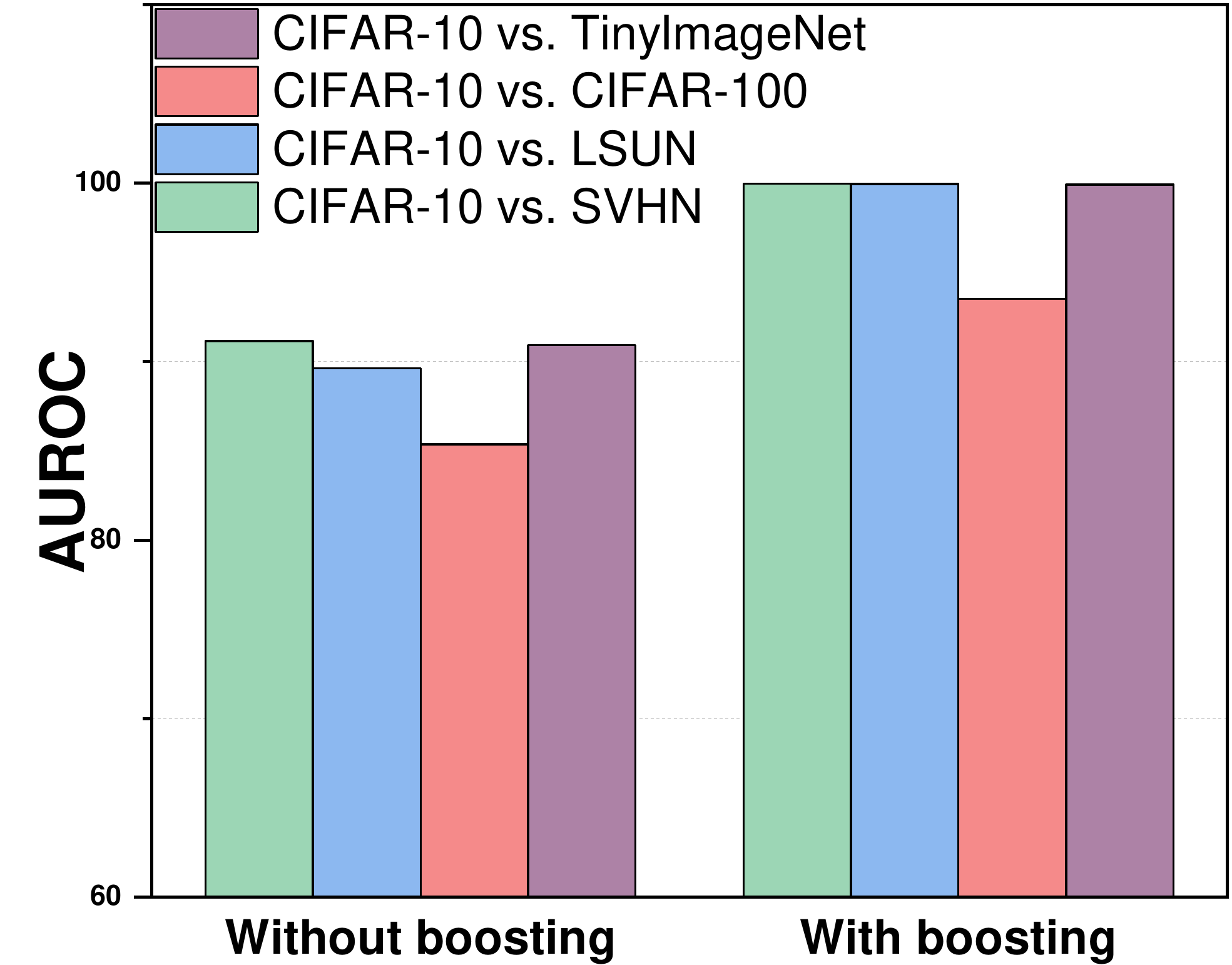}
    \subcaption[]{}
\end{subfigure}\hfill
\begin{subfigure}[t]{0.236\textwidth}
    \centering
    \includegraphics[width=\textwidth]{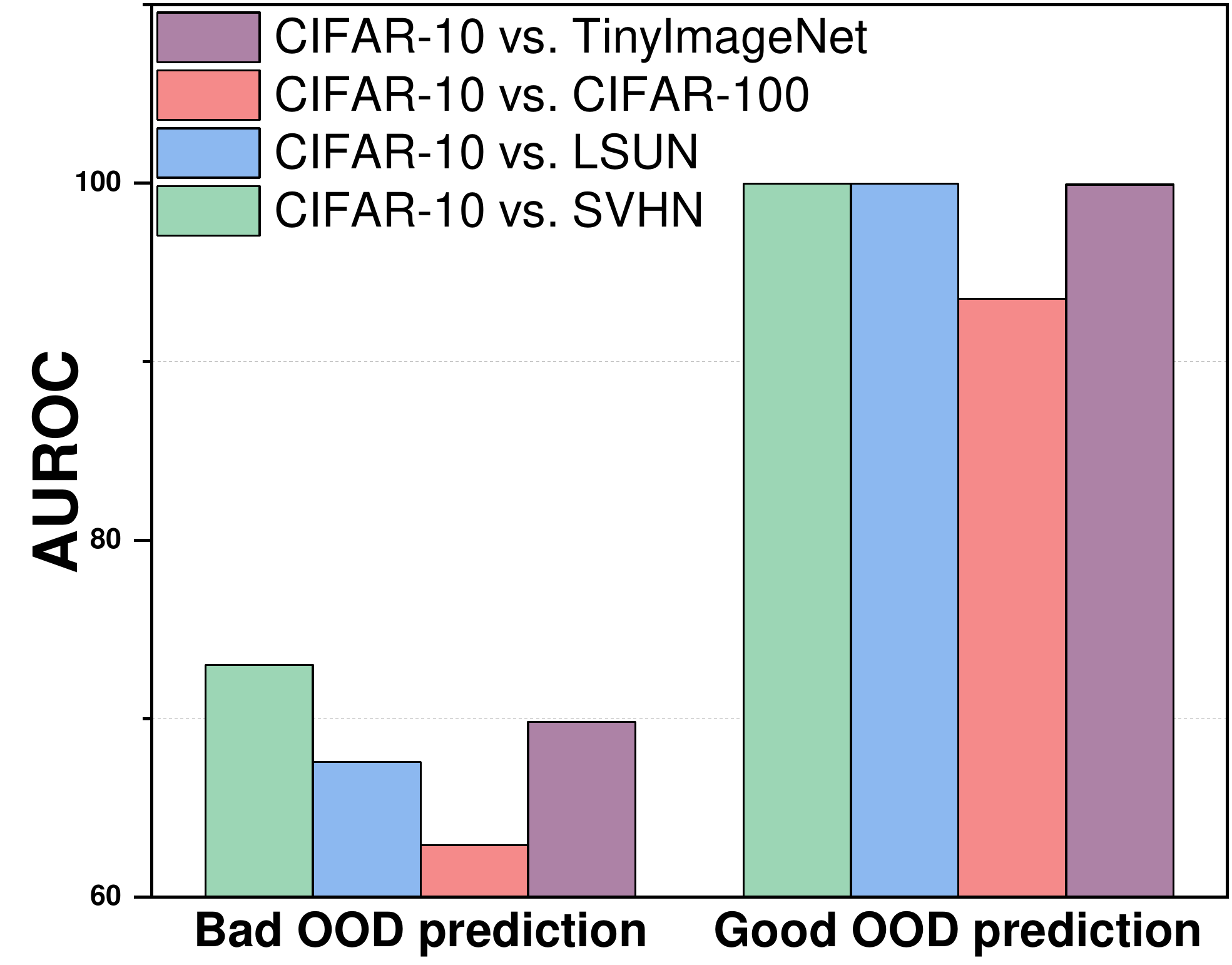}
    \subcaption[]{}
\end{subfigure}
\begin{subfigure}[b]{0.236\textwidth}
    \includegraphics[width=\textwidth]{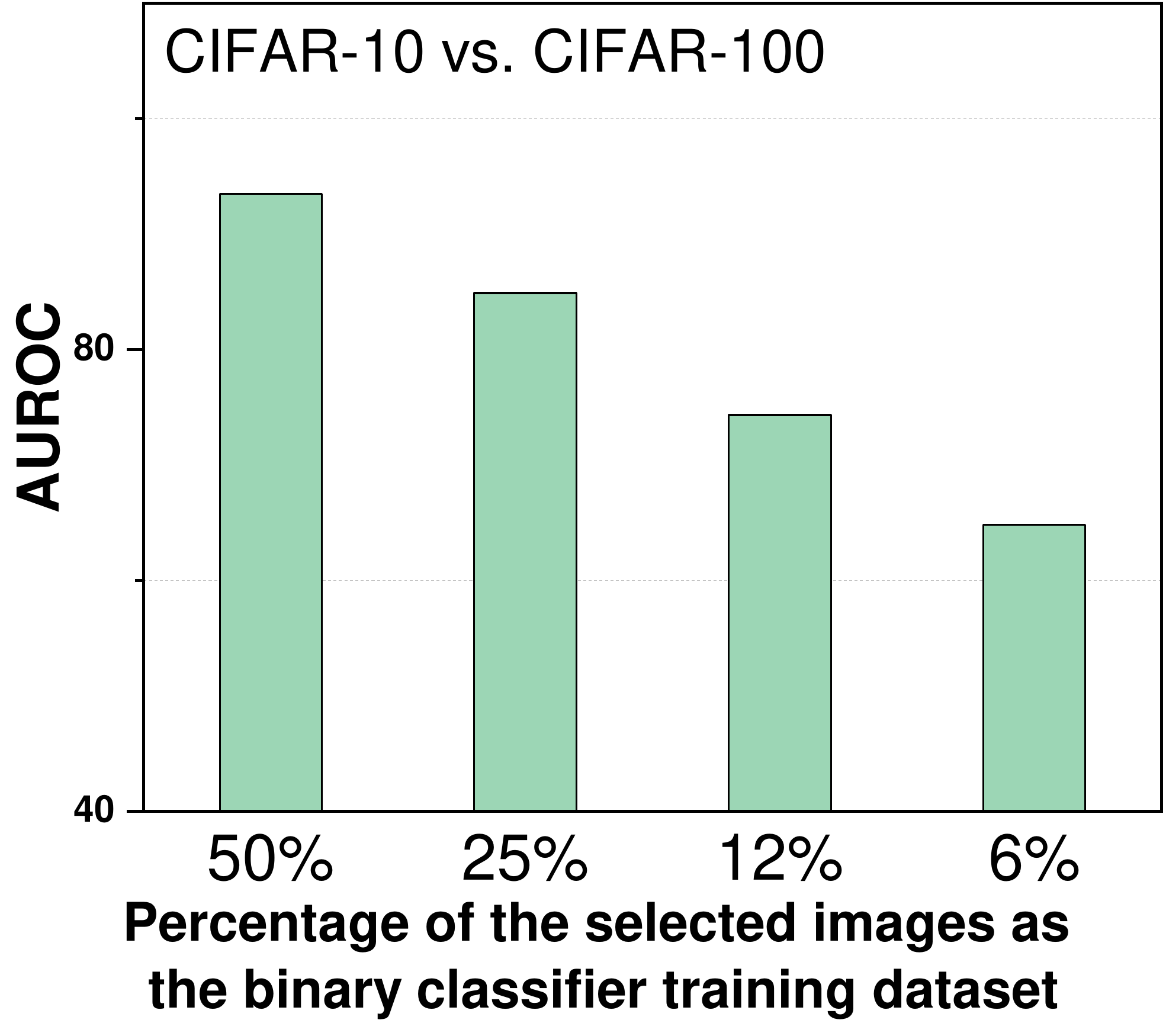}
    \subcaption[]{}
\end{subfigure}\hfill
\begin{subfigure}[b]{0.236\textwidth}
    \includegraphics[width=\textwidth]{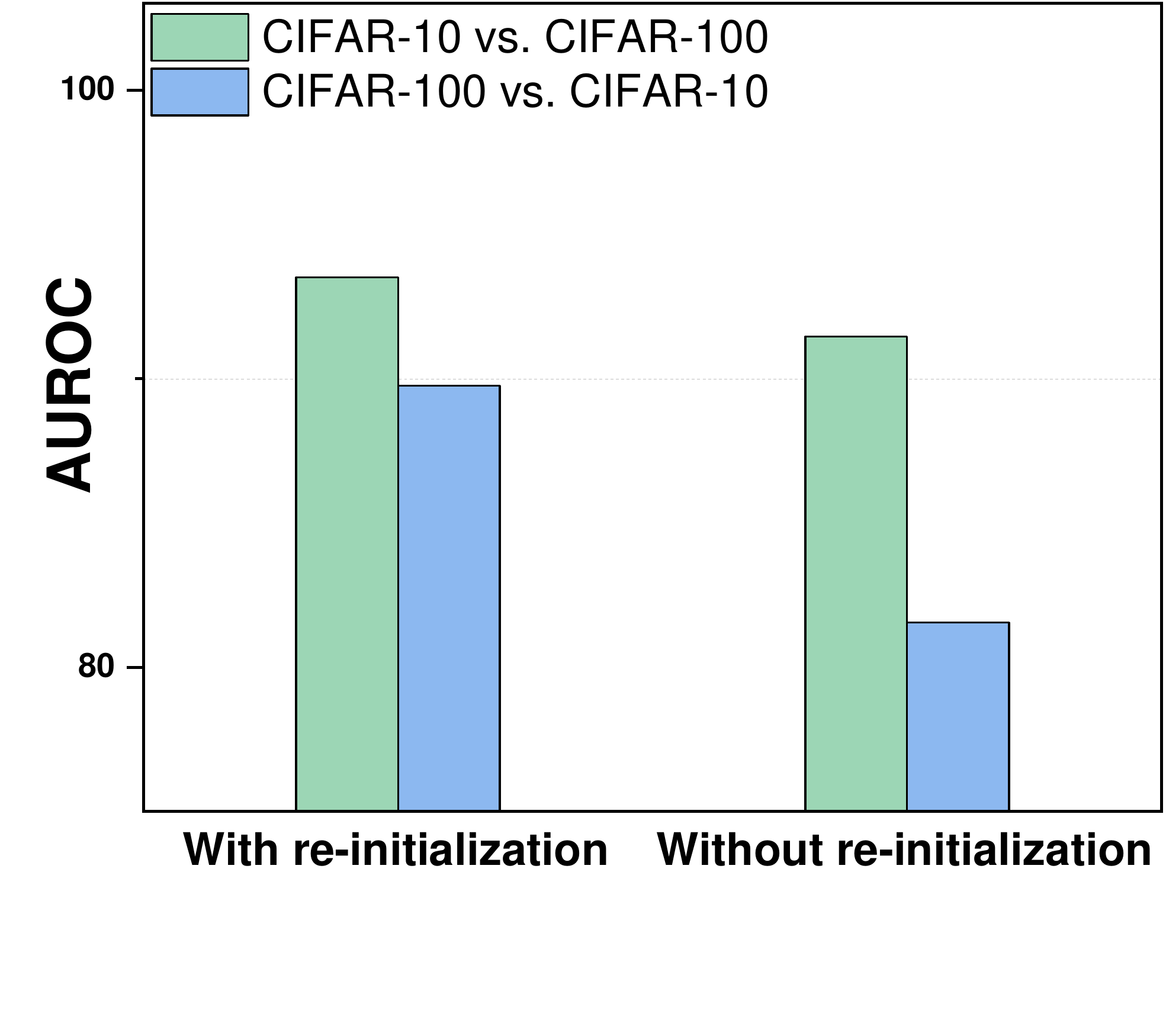}
    \subcaption[]{}
\end{subfigure}

\vspace{-0.2cm}
\caption{(a) \yc{Comparison of} AUROC with and without the binary classifier boosting. (b) \yc{Comparison of} AUROC \yc{between} the good/bad OOD prediction from the novelty detector. (c) AUROC by using different percentages of the selected images to train the binary classifier. (d) \jb{Comparison of AUROC with and without binary classifier re-initialization.}
}
\label{fig:bar}
\end{figure}


\subsection{\ycao{One-class learning with the OOD detector}}

\begin{table}[]

\centering
\resizebox{0.45\textwidth}{!}{
\begin{tabular}{c|cc}
\toprule
CIFAR-10 (9+1) & \multicolumn{2}{c}{Single-head Accuracy} \\ \midrule
New task (OOD) & Memory budget=500 & Memory budget=1000 \\ \midrule
Airplane       & 90.58 & \textbf{95.56}                            \\
Automobile     & 82.16 & \textbf{95.33}                            \\
Bird           & 90.91 & \textbf{91.08}                             \\
Cat            & 91.70 & \textbf{94.54}                              \\
Deer           & \textbf{93.32} & 92.40                             \\
Dog            & 83.28 & \textbf{94.98}                              \\
Frog           & \textbf{96.32} & 96.06                          \\
Horse          & 86.42 & \textbf{93.50}                              \\
Ship           & \textbf{96.48} & 96.44                           \\
Truck          & 84.29 & \textbf{95.39}                              \\ \bottomrule
\end{tabular}
}

\caption{Single-head accuracy tested on each class of CIFAR-10 \ycao{in continual learning. 9+1: 9 classes are first learned together and the last one class is learned through the OOD detector.}} 
\label{continuallearningtest}
\vspace{-0.3cm}
\end{table}

We \ycao{further apply our OOD detector to one-class learning, using CIFAR-10 as the example.} 
\ycao{In this case,} we \ycao{first} train the ResNet-34 network \ycao{with 9 classes (IDD) together}
and \ycao{then} consider the \ycao{last tenth} class as the new learning task (OOD). 
Different from previous experimental setup where \ycao{an} equal amount of IDD/OOD samples (5,000/5,000) are \ycao{used}, we \ycao{attempt to minimize the} 
amount of IDD samples (i.e., 500 and 1,000 in our experiments) to emulate the memory rehearsal. \ycao{We still} allow more OOD samples (2,500) streaming into the system. 
\ycao{In addition to} predict each testing input as either IDD or OOD, we \ycao{further} send all the predicted IDD samples back to the pre-trained \ycao{IDD classifier to recognize the belonging to those 9 IDD classes. Therefore, this procedure completes one-class continual learning. In the field, this solution will temporally help the system to manage incoming OOD data until a new model with updated IDD+OOD classes is available.}

We test the 9+1 single-head accuracy on each class with different memory budgets (Table~\ref{continuallearningtest}). Compared with~\cite{du2020noise} which used memory budget of 2,000 \ycao{IDD samples with the accuracy $<$90\%}, our proposed method achieves \ycao{91-96\% accuracy in any class} 
with only 1,000 \ycao{IDD samples}
and competitive performance even with 500 IDD samples.

%% file: auroctable2.tex
\begin{table*}[!t]
\centering

\resizebox{0.8\textwidth}{!}{
\begin{tabular}{cc|ccccccc}
\toprule  
\multicolumn{2}{c|}{Dataset}                       & \multicolumn{7}{c}{AUROC}               \\ \midrule
\multicolumn{1}{c|}{IDD}            & OOD          & \begin{tabular}[c]{@{}c@{}}Ours\\ (Batch=8)\end{tabular} & \begin{tabular}[c]{@{}c@{}}Ours\\ (Batch=32)\end{tabular} & \begin{tabular}[c]{@{}c@{}}Ours\\ (Batch=128)\end{tabular} & \begin{tabular}[c]{@{}c@{}}SSD$^1$\end{tabular} & \begin{tabular}[c]{@{}c@{}}Mahalanobis$^2$\\(feature based)\end{tabular} & \begin{tabular}[c]{@{}c@{}}Confounding\\label$^3$\end{tabular} & \begin{tabular}[c]{@{}c@{}}ODIN$^4$\end{tabular} \\ \midrule

\multicolumn{1}{c|}{\multirow{4}{*}{CIFAR-10}}  & TinyImageNet & 99.90          & \textbf{99.92} & 99.85           & -    & 99.5       & 93.18     & 98.5 \\
\multicolumn{1}{c|}{}                           & CIFAR-100    & 84.79          & 93.51          & \textbf{97.99}  & 94.0 & -           & -        & -    \\
\multicolumn{1}{c|}{}                           & LSUN         & \textbf{99.99} & 99.97          & \textbf{99.99}  & -    & 99.7        & 99.86    & 99.2    \\
\multicolumn{1}{c|}{}                           & SVHN         & 99.94          & \textbf{99.99} & \textbf{99.99}  & 99.9 & 99.1        & 99.84    & -    \\ \midrule

\multicolumn{1}{c|}{\multirow{4}{*}{CIFAR-100}} & TinyImageNet & \textbf{99.49} & 99.28          & 99.28           & -    & 98.2        & -        & 85.5    \\
\multicolumn{1}{c|}{}                           & CIFAR-10     & 72.61          & 89.76          & \textbf{91.38}  & 84.0 & -           & -        & -    \\
\multicolumn{1}{c|}{}                           & LSUN         & 99.57          & \textbf{99.59} & 99.55           & -    & 98.2        & -        & 86.0    \\
\multicolumn{1}{c|}{}                           & SVHN         & 99.79          & 99.79          & \textbf{99.82}  & 99.5 & 98.4        & -        & -    \\ 
\midrule
\multicolumn{1}{c|}{\multirow{4}{*}{SVHN}}      & CIFAR-10     & 99.94          & \textbf{99.96} & 99.95           & -    & 99.3        & 99.79    & -    \\
\multicolumn{1}{c|}{}                           & TinyImageNet & 99.95          & \textbf{99.97} & 99.95           & -    & 99.9        & 99.77    & -    \\
\multicolumn{1}{c|}{}                           & LSUN         & 99.96          & 99.97          & \textbf{99.98}  & -    & 99.9        & 99.93    & -    \\
\multicolumn{1}{c|}{}                           & CIFAR-100    & 99.65          & 99.74          & \textbf{99.78}  & -    & -           & -        & -    \\ 
\bottomrule 
\multicolumn{9}{l} {\footnotesize $^1$~\protect\cite{sehwag2021ssd}.$^2$~\protect\cite{lee2018simple}.$^3$~\protect\cite{lee2020gradients}.$^4$~\protect\cite{liang2017enhancing}} 
\end{tabular}

}

\caption{Comprehensive evaluation of AUROC on multiple IDD and OOD benchmarks.}
\label{AUPRtable}
\vspace{-0.2cm}
\end{table*}

\begin{table*}[!ht]
\begin{minipage}[t]{0.58\textwidth}
\vspace{-167pt}

\resizebox{\textwidth}{!}{

\begin{tabular}{ cc|c @{\hspace{0.8\tabcolsep}} c @{\hspace{0.8\tabcolsep}} c @{\hspace{0.8\tabcolsep}} c @{\hspace{0.8\tabcolsep}} c}
\toprule
\multicolumn{2}{c|}{Dataset}                       & \multicolumn{5}{c}{AUPR}                                           \\ \midrule
\multicolumn{1}{c|}{IDD}            & OOD          & \begin{tabular}[c]{@{}c@{}}Ours\\ (Batch=8)\end{tabular} & \begin{tabular}[c]{@{}c@{}}Ours\\ (Batch=32)\end{tabular} & \begin{tabular}[c]{@{}c@{}}Ours\\ (Batch=128)\end{tabular} & SSD$^1$  & \begin{tabular}[c]{@{}c@{}}Confounding\\ label$^3$\end{tabular} \\ \midrule
\multicolumn{1}{c|}{\multirow{4}{*}{CIFAR-10}}  & TinyImageNet & 99.91          & \textbf{99.93} & 99.88           & -    & 92.66               \\
\multicolumn{1}{c|}{}                           & CIFAR-100    & 81.46          & 92.21          & \textbf{97.78}  & 92.9 & -                       \\
\multicolumn{1}{c|}{}                           & LSUN         & 99.99          & 99.95          & \textbf{100.0}  & -    & 99.87                \\
\multicolumn{1}{c|}{}                           & SVHN         & 99.98          & 99.99 & \textbf{100.0}           & \textbf{100.0}& 99.98                \\ \midrule
\multicolumn{1}{c|}{\multirow{4}{*}{CIFAR-100}} & TinyImageNet & \textbf{99.43} & 99.22          & 99.23           & -    & -                    \\
\multicolumn{1}{c|}{}                           & CIFAR-10     & 70.07          & 86.99         & \textbf{87.82}   & 81.7 & -                       \\
\multicolumn{1}{c|}{}                           & LSUN         & \textbf{99.53} & 99.52 & 99.51                    & -    & -                    \\
\multicolumn{1}{c|}{}                           & SVHN         & 99.95          & \textbf{99.96} & 99.95           & 99.8 & -                    \\ \midrule
\multicolumn{1}{c|}{\multirow{4}{*}{SVHN}}      & CIFAR-10     & 99.72          & 99.46          & \textbf{99.87}  & -    & 98.11                \\
\multicolumn{1}{c|}{}                           & TinyImageNet & 99.82          & \textbf{99.91} & 99.85           & -    & 97.93                \\
\multicolumn{1}{c|}{}                           & LSUN         & 99.89          & 99.92         & \textbf{99.93}   & -    & 99.21                \\
\multicolumn{1}{c|}{}                           & CIFAR-100    & 99.25          & 99.39         & \textbf{99.49}   & -    & -                      \\ 
\bottomrule
\multicolumn{7}{l} {\footnotesize $^1$~\protect\cite{sehwag2021ssd}.$^3$~\protect\cite{lee2020gradients}} 
\end{tabular}
}

\vspace{9.5pt}
\caption{Comprehensive evaluation of AUPR on multiple IDD and OOD benchmarks.}
\label{aupr}
\end{minipage}
\hfill
\begin{minipage}[t]{0.4\textwidth}
\centering

\includegraphics[width=\textwidth]{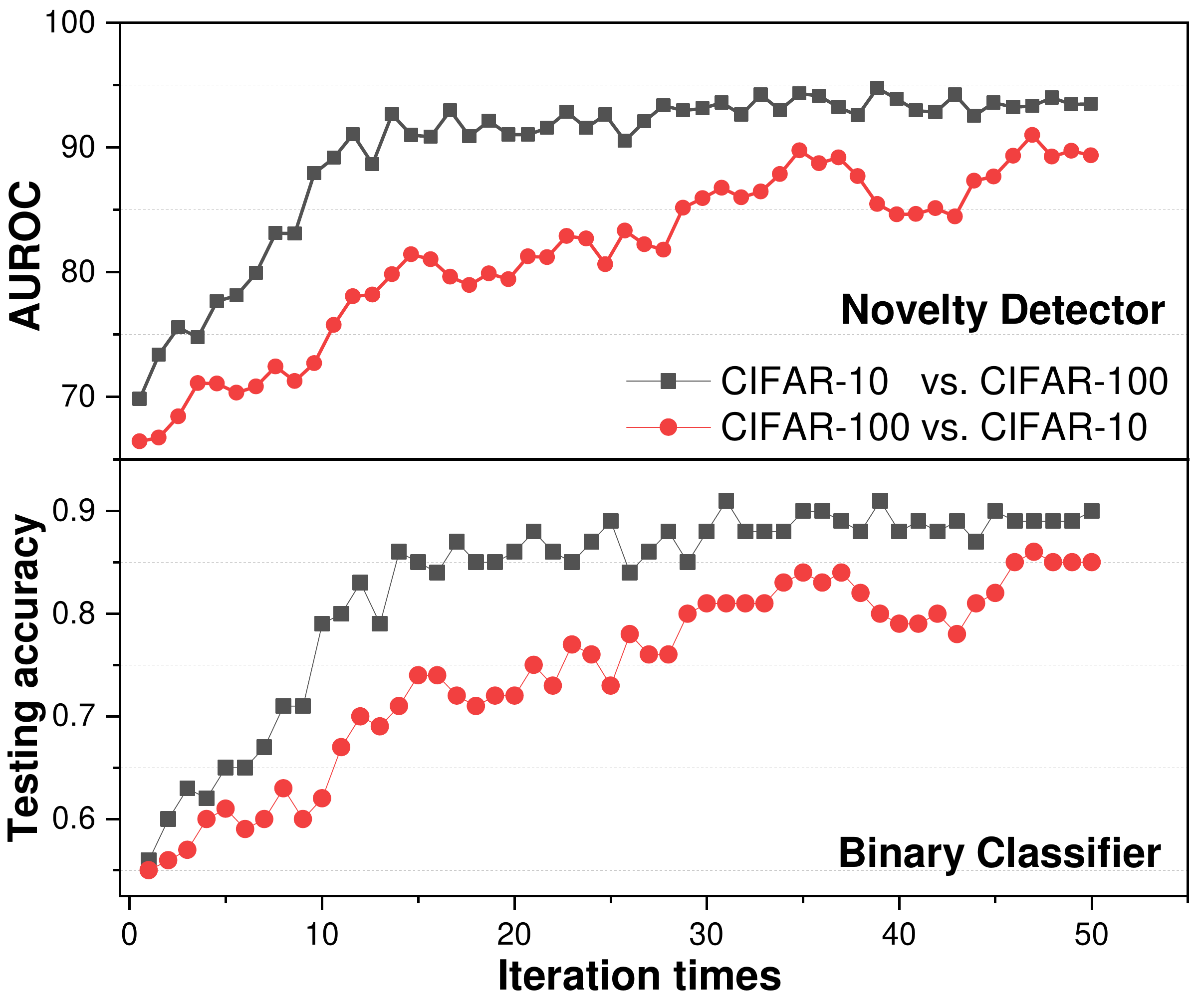}

\captionof{figure}{\yc{AUROC in our novelty detector (top) and the testing accuracy of the binary classifier (bottom) during the mutual assistance (i.e., number of iterations).} }
\label{fig:landscape}
\end{minipage}
\end{table*}

%% file: ablation.tex
\section{Ablation Study and Discussion}

We first analyze the importance of the binary classifier and \yc{the} novelty detector individually
by conducting two ablation studies: (1) With and without the binary classifier, what's the \yc{performance difference of the} gradient-based novelty detector? (2) If the novelty detector \yc{has a poor performance to differentiate} IDD/OOD \yc{and prepare the sample dataset for the training of} the binary classifier, how \yc{does} it influence the overall performance? Fig.~\ref{fig:bar}(a) presents the AUROC results of four OOD test cases. All four test cases benefit from the boosting effect of the binary classifier. To conduct the second ablation study, we randomly select samples from the input batch as the 
training dataset of the binary classifier. This emulates the situation when \yc{the} novelty detector fails to output meaningful novelty scores. As a result, the binary classifier cannot \yc{be} well trained through the predicted IDD/OOD dataset. As shown in Fig.~\ref{fig:bar}(b), a good prediction result from the novelty detector boosts the \yc{inference} accuracy of the binary classifier, which will in return help improve the overall performance. 

\jb{We further conduct two ablation studies regarding the training of the binary classifier: (1) With different percentage of the selected images as the training dataset of the binary classifier, what is the influence on the overall performance? (2) If the binary classifier is not re-initialized after each iteration, will it still converge and provide good guidance for the novelty detection? } 
As shown in Fig.~\ref{fig:bar}(c), when \yc{fewer} IDD/OOD images from the prediction of the novelty detector \yc{are selected} to train the binary classifier, the overall AUROC result decreases. Even though \yc{the selection of fewer data better separates IDD and ODD samples,} 
we also exclude the samples in which IDD and OOD are very similar. Therefore, the binary classifier receives imbalanced training and 
\yc{has a degraded} prediction on the testing dataset.  

\jb{Fig.~\ref{fig:bar}(d) presents the CIFAR-10 vs. CIFAR-100 AUROC results with and without re-training the binary classifier after each iteration. In both experiments, the AUROC decreases when we choose not to re-initialize the binary classifier. This is because during the early stage, the novelty detector is not able to accurately detect the outlier and thus, the predicted training dataset for the binary classifier may suffer from inaccurate IDD/OOD labeling. This ambiguity on IDD/OOD affects the binary classifier in the following iterations during our training process. On the other hand, re-training the binary classifier helps improve the confidence of the novelty detector with more accurate IDD/OOD detection. 
}

%% file: conclusion.tex
\section{Conclusion}

In this paper, we propose a new framework for novelty detection and one-class continual learning,  which involves the \gk{cooperation} of a gradient-based novelty detector and a self-supervised binary classifier with mutually assistance \yc{between these two components}. 
Our proposed framework outperforms previous supervised and unsupervised novelty detectors, as well as one-class continual learning.
\yc{For instance}, our method achieves \yc{superior} performance on CIFAR-100 vs. CIFAR-10 case by improving AUROC from 84\% to 91\% \yc{compared with the state-of-the-art}. 
\ycao{With lower memory budget,} our single-head accuracy of one-class continual learning also outperforms previous methods. 
\ycao{The success of the new approach promises accurate and robust novelty detection in continual learning and other applications.} 

\section*{Acknowledgements}

This research was supported in part by the U.S. Department of Energy, through the Office of Advanced Scientific Computing Research’s “Data-Driven Decision Control for Complex Systems (DnC2S)” project. It was also partially supported by C-BRIC, one of six centers in JUMP, a Semiconductor Research Corporation (SRC) program sponsored by DARPA.
Pacific Northwest National Laboratory is operated by Battelle Memorial Institute for the U.S. Department of Energy under Contract No. DE-AC05-76RL01830. Oak Ridge National Laboratory is operated by UT-Battelle LLC for the U.S. Department of Energy under contract number DE-AC05-00OR22725.